# Dendritic Cells for Anomaly Detection

Julie Greensmith, Jamie Twycross and Uwe Aickelin

*Abstract*— Artificial immune systems, more specifically the negative selection algorithm, have previously been applied to intrusion detection. The aim of this research is to develop an intrusion detection system based on a novel concept in immunology, the Danger Theory. Dendritic Cells (DCs) are antigen presenting cells and key to the activation of the human immune system. DCs perform the vital role of combining signals from the host tissue and correlate these signals with proteins known as antigens. In algorithmic terms, individual DCs perform multi-sensor data fusion based on time-windows. The whole population of DCs asynchronously correlates the fused signals with a secondary data stream. The behaviour of human DCs is abstracted to form the DC Algorithm (DCA), which is implemented using an immune inspired framework, `libtissue`. This system is used to detect context switching for a basic machine learning dataset and to detect outgoing portscans in real-time. Experimental results show a significant difference between an outgoing portscan and normal traffic.

## I. INTRODUCTION

Applications which monitor computer systems and networks for misuse and abuse are known as intrusion detection systems (IDS). Misuse based IDSs use pre-defined paths and patterns of program execution to detect intrusions, but are blind to novel or variant attacks. Anomaly based intrusion detection systems on the other hand generate profiles of 'normal' behaviour [7] [10]. Deviations from normal are classed as 'anomalous' and an alert is generated providing notification of a potential intrusion. Anomaly based systems can produce large amounts of false alarms (false positives), largely caused by the lack of environmental awareness or context. This paper focuses on the development of a contextually aware IDS, built on a foundation of immune inspired components.

A variety of machine learning techniques has been applied to anomaly detection, including neural networks and statistical learning algorithms [7]. Artificial Immune Systems (AIS), based on the functioning of the human immune system, have also been applied to anomaly detection. In 1999, Hofmeyr [10] developed an artificial immune system based on 'negative selection': detectors forming the normal profile are deleted if they match a string denoting normal behaviour. At the time, it was perceived to function in a similar way to the selection of T-lymphocyte cells in the thymus. Problems with negative selection were highlighted by Kim & Bentley [12] and more recently by Stibor et al [16]. Aickelin et al [4] proposed that the negative selection algorithm could not work because it was based on a simplified version of the immunological self-nonself theory. This theory has been challenged within immunology itself and an alternative theory has been proposed - the Danger Theory [13]. The Danger Theory states that the immune system does not discriminate on the basis of self or nonself, but on the balance between the concentration of danger and 'safe' signals within the tissue of the body. The work described in this paper is the application of danger theory ideas to anomaly based intrusion detection.

It is our hypothesis that the inclusion of more biologically realistic features alleviates some of the scaling problems and reduces the amount of false alerts generated by negative selection based systems. In this paper we propose and test an algorithm based on the behaviour of DCs, who are major players in the danger theory. DCs are responsible for combining signals in the tissue and informing the immune system of any changes in signal concentration. We have used an abstraction of DC behaviour to develop the DCA. In this paper, we demonstrate the application of the DCA to two problems - a static machine learning problem and a computer security related real-time problem. This work is an extension of concepts and experiments described in [8]. We verify the correctness of the algorithm's implementation, and present experiments demonstrating real-time anomaly detection properties of the DCA. Further experiments are required to demonstrate the DCA's effectiveness as an an anomaly detector in other domains.

In this paper, Section II contains background information describing the biological basis for the new algorithm and information on the development of AIS for IDS; Section III briefly describes the DCA; Section IV outlines experiments performed on a static dataset, complete with results; Section V contains experimental details from testing the algorithm in a pre-intrusion scenario; Section VI contains results and analysis of the intrusion experiment; and finally, Section VII provides conclusions and future directions.

## II. BACKGROUND

In order to describe the algorithm it is first necessary to explain some of the biological concepts used and to place these concepts in context within the field of artificial immune systems. The biological basis for the functioning of DCs has been described in detail in our previous work [8]. Subsequent sections explain the relevance of the danger theory to the DCA and how this can be applied to intrusion detection.

### A. The Rise and Fall of Self-Nonself

Until relatively recently the human immune system was thought to discriminate between antigens (proteins) belonging to 'self' versus antigens belonging to pathogens - 'non-self'. The self-nonself theory relies on the expression of antigen-recognising receptors present on T-lymphocytes.

School of Computer Science, University of Nottingham, NG8 1BB, UK. E-mail {jqg, jpt, uxa}@cs.nott.ac.uk

These receptors are matched against self antigen during a training period in the thymus, resulting in the deletion of cells with the potential to interact with self antigens. Recently, several questions have been raised regarding the validity of this model as a central tenet. What is defined as 'self' actually changes throughout the lifetime of an individual, e.g. a pregnant woman's immune system does not react against her unborn foetus despite consisting of 'nonself' proteins. A modification to the self-nonself theory was proposed by Janeway [11], namely the 'infectious non-self' model. This states that an antigen must be associated with a PAMP (pathogen associated molecular patterns) in order to trigger a response, as recognised by the innate immune system. While this model could explain the rationale for adding stimulatory adjuvants to vaccines, it could still not answer pertinent questions relating to autoimmunity (e.g. Multiple Sclerosis).

The Danger Theory [13] provides an alternative view of the activation of the immune system. Unlike the detection of non-self antigens or pathogenic molecules, the danger model proposes that the immune system detects the presence of danger signals, released as a result of necrotic cell death within the host tissue. Necrosis is the result of cellular damage and stress caused by pathogenic infection or exposure to extreme conditions. The metabolites of internal cell components are thought to form the danger signals and are released into the surround buffer fluid. The cell membrane loses its integrity, releasing its contents (e.g. DNA, mitochondria) into the surrounding tissue fluid [14]. The Danger Theory proposes that the immune system is sensitive to changes in the danger signal concentration in the tissue. Conversely, when then tissue is healthy, cells die in a controlled manner, known as apoptosis. Immunosuppressive molecules (safe signals) are released as an indicator of normality in the tissue. In essence, the Danger Theory consists of active suppression while the tissue is healthy (apopotosis), combined with rapid activation on receipt of necrotic danger signals. This property can be abstracted to form artificial tissue, as conceptualised in a software framework in [17] or used for data representation in [6].

DC's first function is to reside in tissue, where they are classed as 'immature'. Whilst in tissue, DCs collect antigen (regardless of the source) and experience danger signals from necrosing cells and 'safe' signals from apoptotic cells. Maturation of DCs occurs in response to the receipt of these signals. On maturation, DC exhibit the following behaviour: collection of antigen ceases; expression of co-stimulatory molecules (necessary for binding to powerful T-lymphocytes) and chemical messengers known as cytokines; migration from the tissue to a lymphatic organ such as a lymph node; and presenting antigen to T-lymphocytes. The context of the tissue (i.e. the type of signals experienced) is reflected in the output chemicals of the DC. If there is a greater concentration of danger signals in the tissue at the time of antigen collection, the DC will become fully mature (mDC), and will express mDC cytokines. Conversely, if the DC is exposed to 'safe' signals, the cell matures differently becoming a semi-mature DC, expressing smDC cytokines [15]. The mDC cytokines activate T-lymphocytes expressing complimentary receptors to the presented antigen. Any peripheral cells expressing that antigen type are removed through the activated T-lymphocyte. The smDC cytokines suppress the activity of any matching T-cell, inducing tolerance to the presented antigen. The context of the antigen is assessed based on the resulting cytokine expression of the DC. In our model, a combination of Janeway's Infectious Non-self (PAMPs) model with Matzinger's Danger model (signals) is used to investigate an artificial DC algorithm.

### B. AIS: The Story So Far

Similarly to immunology, AIS initially relied on self-nonself principles to create algorithms such as negative selection and the B-cell based clonal expansion. Negative selection has been used extensively for the purpose of intrusion detection [5], with numerous variations for antigen representation. The algorithm described in [10] is used to generate a normal profile based on a detector set. In their work, network connections established by host machines are monitored and mapped onto a schema for matching. Their model uses a training period in which common connections are deleted from the detector set based on an activation threshold. Once the training period is finished, the detectors are compared against new connections. Connections with a sufficiently high match count to a particular detector are classed as an anomaly and an alert is generated. This algorithm has been criticised for problems with false positives and scaling. False positives arise because the nature of a machine's connections can change over time as part of normal behaviour. As the detectors are not dynamically re-trained, once a connection rate threshold is set, any change in the behaviour of the machine results in the AIS responding to a seemingly anomalous connection. An additional problem is detector generation. Initially, the values for the detectors are created at random. As the size of the connection space increases, the size of the random detector set grows at an unsustainable rate. This criticism has been theoretically validated by the work of Stibor et al [16].

While this algorithm advanced the development of AIS by stimulating further work in this field, it also hindered development by inspiring other algorithms to be based on similar, simplistic views of immunology. The Danger Project, initially proposed in 2003, aims to improve on the results of negative selection based IDSs by re-thinking the use of immunology within AIS. The work presented in this paper is the implementation of ideas outlined in this proposal [4].

### III. THE DC ALGORITHM AND libtissue

#### A. The DC Algorithm

DCs have a number of functional properties that can be abstracted to form a useful algorithm. The abstraction process, algorithm and a worked example of the signal processing methods are described in detail in our previous work [8]. Outlined below are the key characteristics of DC behaviour used to generate the DCA:

- Immature DCs (iDC) collect multiple antigens and are exposed to signals in the tissue.
- DCs can combine signals from multiple sources to generate different output concentrations of costimulatory molecules, semi-mature cytokines and mature cytokines.
- Exposure to signals generates an increase in co-stimulatory molecules, with a high amount leading to migration to the lymph node.
- Exposure to signals causes the maturation of DCs into either mature and semi mature states, assessed through the differences in concentration of mature and semi-mature cytokines.

For the purpose of this model, a simple interpretation of the input signals has been derived. There are four signals in our model, each from a different source and producing different output cytokines:

- PAMPS (P) are based on pre-defined signatures. Exposure to PAMPS causes an increase in mDC cytokines. PAMPs are suppressed by safe signals.
- Danger signals (D) cause an increase in mDC cytokines. Danger Signals can also be suppressed by safe signals. Danger signals have a lower potency than PAMPs.
- Safe signals (S) cause an increase in smDC cytokines and have a suppressive effect on both PAMPS and danger signals.
- Inflammatory cytokines (IC) amplify the effects of the other three signals, but are not sufficient to cause any effect on DCs when used in isolation.

Our data and method of processing is very different from other AIS, which rely on pattern matching of antigen to drive their systems, e.g. [19]. In our algorithm, antigen is only used for the labelling and tracking of data, hence we do not have a similarity metric. The representation of the antigen can be a string of either integers or characters. Signals are represented as real-valued numbers, proportional to values derived from the context information of the dataset in use. For example, a danger signal may be an increase in CPU usage of a computer. The value for the CPU load can be normalised within a range and converted into its real-valued signal concentration value. The signal values are combined using a weighted function (Equation 1) with suggested values of the weights derived from empirical data based on immunologists' wet lab results (Dr Julie McLeod, Dr Rachel Harry and Charlotte Williams - University of the West of England).

The function itself is a weighted sum of PAMP, danger and safe signal concentration values, multiplied by a value for inflammation (in the range of 0 and 2). The resulting value is then normalised through division by the sum of the weights. The function is used three times to calculate the output cytokines of costimulatory molecules (CSMs), mDC and smDC cytokines, as denoted. C represents a concentration, with the subscript denoting the cytokine or signal it represents (P, D, S, IC are PAMP, danger, safe and inflammatory signals). Similarly, W corresponds to the weights used. These values are updated each time a DC is exposed to signals, which influence the state of the DC: immature or mature, with immature being the initial, default state for the cells. Transition to the mature state is facilitated through the CSM value. Each cell is assigned an individual migration CSM threshold value, which can vary between cells should it be required. When a cell's CSM value exceeds the migration threshold, the status of the cell changes and migration from the tissue is initiated.

$$C_{[csm,mDCi,smDC]} = \frac{((W_P * C_P)+(W_S * C_S)+(W_D * C_D))}{(W_P+W_S+W_D)} * \frac{1+IC}{2}$$

(1)

TABLE I
WEIGHTS FOR THE SIGNAL PROCESSING FUNCTION

| W | csm | semi | mat |
|---|---|---|---|
| PAMPs(P) | 2 | 0 | 2 |
| Danger Signals(D) | 1 | 0 | 1 |
| Safe Signals (S) | 2 | 3 | -3 |

The DCA is a population based algorithm, with a user defined number of DCs created to form a sampling pool. While in the sampling pool, each DC is exposed to current signal values and selects a slot in the antigen store. If an antigen is present in the antigen store, the DC collects the antigen and ingests it in the DC internal antigen storage. Each DC has the opportunity to sample multiple antigens. For every iteration of antigen collection, each DC re-calculates its internal cytokine values based on the input signals received. Each antigen can be sampled single or multiple times (a tunable parameter).

Migration is simulated by the removal of a DC from the pool and occurs when the cell's internal CSM value exceeds the DC's migration threshold. At this point, the output cytokines of each DC are measured. Antigen presented by cells predominantly expressing mature cytokines is labelled 'mature context antigen', whilst antigen from cells expressing predominantly semi-mature cytokines is labelled as 'semi-mature'. Each presented antigen's context is recorded and eventually a mean antigen context value (between 0 and 1) is derived. Further details of the DCA can be found in [8].

B. System Integration

`libtissue` [17] is a software system which allows the implementation and testing of AIS algorithms on real-world computer security problems. Its design comes from detailed research into innate immunology [18] and computer security. It allows researchers to implement AIS algorithms as a collection of cells, antigen and signals, interacting within a tissue compartment. Input data to the tissue compartment comes in the form of real-time events generated by sensors monitoring a system under surveillance. Cells are actively able to affect the monitored system through response mechanisms, though none are relevant to the DCA.

The system has a client/server architecture pictured in Figure 1. The AIS algorithm is implemented as a `libtissue`

server, while `libtissue` clients provide input data to the algorithm and provide response mechanisms. This client/server architecture separates data collection by the `libtissue` clients from data processing by the `libtissue` servers. Client and server APIs exist, allowing new antigen and signal sources to be easily added to `libtissue` servers, and the testing of the same algorithm with a number of different data sources. Client/server communication is socket-based and uses the SCTP protocol, allowing clients and servers to potentially run on separate machines, for example a signal client may in fact be a remote network monitor.

`libtissue` clients are of three types: antigen, signal and response. Antigen clients collect data and transform it into antigen which are forwarded to a `libtissue` server. Currently, a systrace antigen client has been implemented which collects process system calls using systrace [3]. System calls are a low-level mechanism by which applications request system services such as peripheral I/O or memory allocation from an operating system. Signal clients monitor system behaviour and provide the AIS with input signals. A process signal client, which monitors a process and its children and records statistics such as CPU and memory usage, and a network signal client, which monitors network interface statistics such as bytes per second, have also been implemented. Currently, response clients which directly modify a systrace system call policy and generate alerts are also implemented. These clients are designed to be used in real-time experiments and for data collection for offline experiments.

The implementation is designed to allow varied AIS algorithms to be evaluated on real-world systems and problems[17]. When testing IDSs it is common to use preexisting datasets such as the Lincoln Labs dataset [1]. However, our project is focused on combining measurements from a number of different concurrent data sources. Pre-existing data sets are not available containing the necessary data for our system. To facilitate experimentation, a replay client has also been implemented. This client reads in log files gathered from previous real-time runs of antigen and signal clients. It also has the facility to read logfiles generated by strace [2] as an optional source of antigen in place of the systrace client. It then sends these logs to a `libtissue` server. Variable replay rates are available, allowing data collected from a real-time session to be used to perform many experiments quickly.

The `libtissue` server itself provides a programming environment in which AIS algorithms can be implemented. Input data for these algorithms comes from data sources provided by connected `libtissue` clients and is represented in a tissue compartment. A tissue compartment is a space in which cells, signals and antigen interact. Each tissue compartment has a fixed-size antigen store where antigen provided by `libtissue` clients is placed. The tissue compartment also stores levels of signals, set either by signal tissue clients or cells.

`libtissue` cells, like tissue compartments, have antigen and signal stores. They also have a number of different receptors and producers which allow cells to interact with others cells, antigen and signals in the tissue compartment. Currently, four types of receptors have been implemented: antigen, cytokine, cell and vr receptors. Antigen receptors allow cells to transfer antigen from the tissue compartment to their own internal antigen store. Cytokine receptor allow cells to read signal levels in the compartment. Cell receptors allow cells to bind to other cells. Binding is necessary for vr receptors, which match antigen presented on another cell, to be activated. Antigen from a cell's internal store are presented on antigen producers, one of the three types of producers currently implemented. The other two types, cytokine and response producers, allow cells to change cytokine levels in the tissue compartment and communicate with response clients respectively.

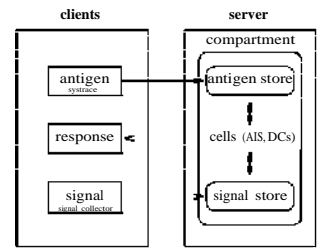

Fig. 1. `libtissue` System Architecture

## IV. WISCONSIN BREAST CANCER DATA

Previous work with the DCA included preliminary experiments using the Wisconsin Breast Cancer dataset [9] to form the signals and antigen, and produced class switching behaviour in the DCs. Here we repeat and extend these experiments to verify the implementation of the DCA within the `libtissue` framework [8].

The Wisconsin Breast Cancer dataset [9] consists of 700 items: 240 items in class 0, 460 in class 1 and with each data item containing nine normalised attributes. The static nature of the dataset enables testing for detection accuracy and provides two different data contexts which can be 'replayed'. Although we are aware that using static data to test an intrusion detection algorithm is not ideal, it nevertheless provides a predictable test-bed for examining the algorithm itself. The antigen context output of the DCA classifies the data into class 0 or class 1. This value is then compared against the original classification and the number of errors recorded.

### A. Signals and Antigen

In our experiments, five out of the nine attributes are used to form the signals. The five attributes with the largest standard deviation are chosen. Cell shape, bare nuclei and normal nucleoli are danger signals. Clump size has the highest standard deviation out of all attributes and is used to calculate PAMP and safe signal values for each data item.

The PAMP and safe signal value equals the deviation from the mean of the class [8].

The signal concentration values are accumulated and processed within each DC using the weighted signal processing function described in Section III. Weightings used for this experiment are derived from empirical immunology and are shown in Table 1. The total signal values form the basis of discrimination for the detection of context changes in the original data.

In this experiment we have not included values for inflammatory cytokines, as no suitable mapping exists. Antigen provides a label for each data item. A migration threshold for removal from the sampling pool is implemented. Once this threshold is exceeded, a DC is removed from the sampling pool and its antigen plus context is written in a log file. Once all the data items are processed the log file is analysed. The number of 'mature' and 'semi-mature' presentations per antigen is calculated. Ultimately, a mean value is calculated from the context data. A threshold reflecting the distribution of the original data is used to discern between antigens presented in a class 0 or class 1 context. Values exceeding the threshold are categorised as class 1.

### B. Tissue Clients and Parameters

In this experiment, a tissue client is used to convert the dataset from a text file into a data-stream structure. These values are represented as signal concentrations and are passed to the tissue server. Parameters are chosen to provide similar conditions as used in previous work [8]. A summary of the parameters used is shown in Table II.

TABLE II
TISSUE SERVER PARAMETER SETTINGS

| Parameter | Breast Cancer Expts | Port Scan Expts |
|---|---|---|
| Tissue antigen capacity | 1 antigen | 500 antigen |
| Cell antigen capacity Number | 50 antigen | 50 antigen |
| of DCs in pool Antigen | 100 DCs | 500 DCs |
| sampling probability Signal | 0.10 | 1.00 |
| decay rate | 100% | 100% |
| Cell cycle rate | 1 sample/s | 1 sample/s |
| Number of times a single antigen is sampled | 10 | 1 |

### C. Experiments

Two series of experiments are performed using the cancer dataset, with all experiments performed on a Debian Linux machine (kernel 2.4.10, AMD Athlon 1GHz). All code is implemented in C (gcc 4.0.2). Each experiment is performed 20 times, with 7000 antigen collected per run. The context values from each run are used to derive an overall mean context value for each antigen. A threshold of 0.65 (reflective of the class distribution of the dataset) is used to determine the final class label of each antigen. Antigen with context values of over 0.65 are designated class 1 and vice versa.

Experimental series 1 explores the class-switching behaviour of the DCs by investigating the influence of the data order. Three data orders are explored: class 0 followed by class 1 (one-step); the first half of class 0, all of class 1, followed by the remainder of class 0 (two-step); and a random distribution. A further experiment is performed using the two-step data order where each antigen is sampled only once, not 10 times as in previous experiments. The migration threshold for each DC is generated randomly within the range of 5-15 (mean value of 10).

The series 2 experiments use the one-step data order to assess the effect of changing the migration threshold value. The migration threshold value limits the amount of antigen a DC can collect. A larger migration threshold increases the sampling window size for antigen collection. We are interested to see what effect variations of this value have on the amount of errors produced by the system. The migration threshold is fixed at 1, 5, 10, 15 and compared to the previously used random number within the range of 5-15. Each experiment is run 20 times with the context values for each antigen derived from the mean of all runs.

### D. The Results

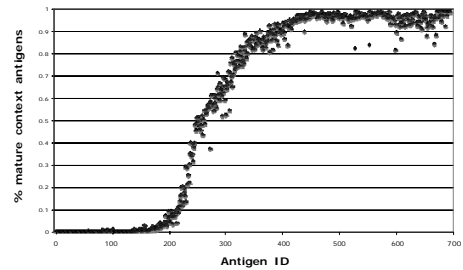

Fig. 2. The experiment shown features a one step data order (n=20). The data set contains 240 data items of class 0 followed by 460 class 1 items. The context switching of DCs is demonstrated in accordance with change in the data.

The results from series 1 are shown in Figure 2 and in Table III. Figure 2 shows the class switching behaviour of the DCs from class 0 to class 1 in line with the data order. These results produce comparable trends with previous results[8]. However, there is some loss of accuracy as seen from the increased amount of errors listed in Table III. This is due to the addition of new components in `libtissue`, such as antigen overwriting and random selection of cells in the update cycle of the DCs. When each antigen is sampled once, the results for the 2-step experiment improve as shown in Table III. This is due to the strong linkage between antigen and signals in this experimental set-up. As expected when the data order is completely random the performance of the algorithm is poor, due to the lack of linkage between antigen and signals.

TABLE III
NUMBER OF ERRORS FOR EACH DATA ORDER

| Data Order | Number of Errors |
|---|---|
| 1-step | 57 |
| 2-step | 133 |
| Random | 349 |
| 2-step (single Ag sample) | 4 |

The results from series 2 are presented in Table IV. The results show that variation in the migration threshold significantly alters the classification of data items (using a paired t-test, $p < 0.01$) around the class-switch boundary. Analysis based on the number of errors suggests that CSM=1 is the best value for the threshold. However, the magnitude of the errors generated using CSM=1 (error values not reported) is greater than those reported when CSM=5, making CSM=5 a more suitable value. The variable threshold was expected to perform within the same range as CSM=10, being the mean of the random range. However, the number of errors produced using CSM=variable was smaller than expected and significantly different to the results from CSM=10. The variable threshold adds robustness to the system; furthermore a set value need not be specified, as long as the values fall within a sensible range.

TABLE IV
NUMBER OF ERRORS FOR EACH DATA ORDER

| Migration Threshold | Number of Errors |
|---|---|
| 1 | 14 |
| 5 | 41 |
| 10 | 83 |
| 15 | 182 |
| Variable | 57 |

## V. PORT SCAN EXPERIMENTS

Port scanning is used primarily for network administration and consists of sending packets to machines for the purpose of understanding the topography of a network. However, this tool is often used for malicious purposes to search out vulnerable machines. Detection of port scans is a key step in the detection of an attack and makes an ideal small scale, real-time experiment for testing the DCA. An essential purpose of this experiment is to ensure the tissue server, clients and algorithm can cope with real-time data processing.

### A. The ICMP scan

The port scan used is a ping (ICMP) scan performed across a range of IP-addresses, using the nmap tool. In order to perform this port scan, a remote shell (via ssh on a selected port, 2222) is established, from which nmap is executed, providing an example of anomalous behaviour. To capture normal behaviour, a file transfer is performed from the host machine to another, in addition to ssh demon processes and shells running in the background. The port scan and the file transfer are not performed simultaneously, to allow for a baseline of comparison. Four different experiments are performed varying the combinations of the different signals and the safe signal weighting for the mature cytokine value. The protocol describing the port scan procedure is as follows:

- Remotely log-in to host using ssh port 2222
- Run the nmap (nmap -sP), ICMP scanning over 1000 IP addresses
- Wait for 30 seconds
- Transfer a file from the host machine (3.3MB)
- Close the remote session

### B. Signals and Antigen

Process IDs from all processes running through the remote shell are captured using a systrace tissue client, as described in Section III. The process IDs are the antigen of the system and are classed as either 'semi-mature' or 'mature' dependent on the percentage of DCs expressing an antigen with mature cytokines. The proportion of mature context antigen (per process) are used to identify misbehaving processes. This context is generated by exposing the DCs to various signals representative of machine behaviour. Signals have been selected to reflect a network based attack. An 'ideal' PAMP would be a signature or definite indicator of a pathogenic presence. In this instance the number of ICMP 'destination unreachable' errors are converted into PAMP signals. These signals are generated if a scanned IP address does not have an associated machine - a potential sign of port scanning.

The amount of packets transmitted per second is measured and forms the danger signals. The higher the amount of packets per second, the greater the concentration of the danger signal. The safe signals are viewed as the opposite of danger signals. Safe signals are derived from the inverse rate of change of packets per second, based on a moving average over a 2 second time window.

Inflammatory signals are reflective of the machine behaviour in general, but not specific to the scan. In this instance, the inflammatory signal was set to zero to indicate 'user present' or set to 1 to simulate 'user absent'. This is based on the assumption that it is less likely to comply as normal data should a port scan take place without a user present at the machine. An example of the incoming signals to the tissue server is shown in Figure 3.

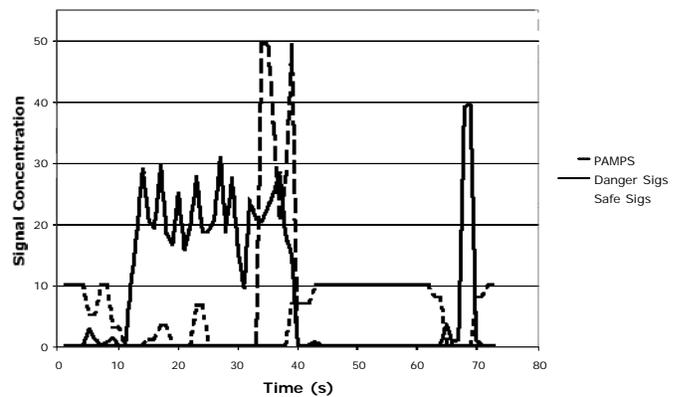

Fig. 3. The PAMPs, danger signal and safe signal inputs.

### C. Parameters, Settings and Scripts

The parameters used in this experiment are shown in Table II. All experiments were performed using nmap version 3.2.1 on an AMD Athlon machine (1GHZ) running Debian Linux kernel version 2.6.10. OpenSSH version 4.2 was used for the remote shell, with all code written in C (gcc version 4.0.2). Signal collection scripts are written in bash to extract relevant data from the /proc file system, in addition to the use of the netstat program. These samples were taken once per second. Each experiment was performed ten times, with the mean percentage of mature antigen calculated to assess the anomalous or normal nature of the individual processes. A process monitor script is used to capture which processes have run throughout the course of the experiment for the purpose of antigen identification. Forked or child process IDs are also captured.

Experiments with the port-scan tissue clients are performed using different combinations of signals. A summary of the experiments is detailed in Table V. As PAMPs did not feature in experiment 1, the weighting for the transformation of safe signals to CSMs was changed from -3 to -1. In these experiments all other weighting values are as described in Table I. This weighting was increased to -2 for experiments 3 and 4 to explore the relationship between detection of normal processes and an increase in safe signal suppression. In experiment 4, the inflammatory signal was used for the first time, to assess the usefulness of the signal. A summary of the experiments performed is detailed in Table V.

TABLE V
PORT SCAN EXPERIMENTS

| Experiment | Signal Types | Safe Signal to CSM Weight |
|---|---|---|
| 1 | danger & safe | -1 |
| 2 | danger, safe & PAMP | -1 |
| 3 | danger, safe & PAMP | -2 |
| 4 | danger, safe, PAMP & inflammation | -2 |

## VI. PORT SCAN RESULTS AND ANALYSIS

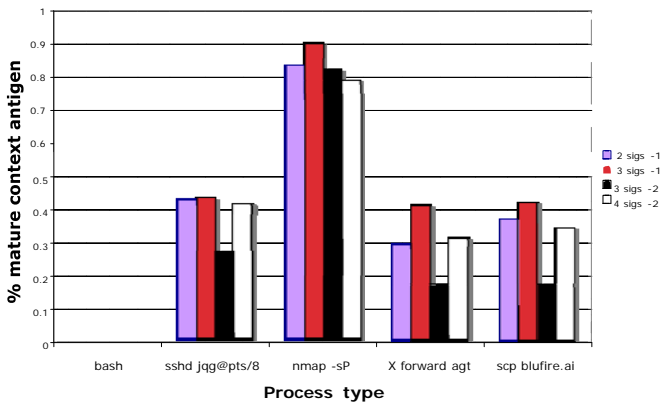

Fig. 4. The mean % mature antigen derived for each process of interest, shown for each signal combination.

In experiment 1, all process IDs presented as antigen are analysed and a mean percentage mature antigen value (% mAg) is calculated. The results in Table VI show that although antigen is presented for numerous sshd processes, negligible amounts are detected as dangerous. Similar values are found for experiments 2, 3 and 4 and thus the sshd child processes are omitted from the corresponding results

TABLE VI
TWO SIGNALS (EXPT 1)

| Process Name | Num antigen | Mean % mAg | Std Dev |
|---|---|---|---|
| ssh script | 10.4 | 0 | 0 |
| sshd -p 2222 | 31.9 | 0 | 0 |
| sshd:jqg[priv] | 246.2 | 0 | 0.002 |
| sshd:jqg[net] | 26.7 | 0 | 0 |
| sshd:jqg[pam] | 10.4 | 0 | 0 |
| bash | 293.6 | 0.205 | 0.159 |
| sshd jqg@pts/8 | 476.0 | 0.437 | 0.090 |
| nmap -sP | 3536.4 | 0.843 | 0.069 |
| X forward agt | 471.7 | 0.297 | 0.115 |
| scp blufire.ai | 285.3 | 0.378 | 0.157 |

TABLE VII
THREE SIGNALS WITH -1 SAFE SIGNAL WEIGHTING (EXPT 2)

| Process Name | Mean % mAg | Standard Deviation |
|---|---|---|
| bash | 0.212 | 0.152 |
| sshd jqg@pts/8 | 0.441 | 0.159 |
| nmap -sP | 0.910 | 0.0456 |
| X forward agt | 0.421 | 0.176 |
| scp blufire.ai | 0.428 | 0.187 |

tables, Tables VII, VIII and IX respectively. The tables show the mean percentage mature antigen for each process in addition to the standard deviation from the mean, found to be reflective of the actual data distribution. The five processes of interest presented in Figure 4 include: the bash shell from which the scan was performed; the ssh demon; the nmap performing the port scan; the graphical forwarding agent for the remote shell; and the file transfer (scp). The scp file transfer provides us with a baseline comparison. If the DCA works correctly, nmap is expected to produce a significantly greater mean % mature antigen than the file transfer. Indeed, analysis of data from all experiments shows that nmap produces significantly greater mean % mature antigen than the baseline normal file transfer (see Table X), especially in experiment 3. This is true for any combination of signals. Furthermore, the mean % mature antigen for the normal file transfer is significantly reduced when the safe signal weight is changed to -2. All significance are assessed through paired t-tests, with 95% confidence demonstrated.

### A. Analysis

In each experiment the nmap process generates significantly more mature context antigen than any other process, as shown through the paired t-test results given in Table X. The standard deviations of these results are within an acceptable range indicating that all means are representative of the sample. The good detection rate of the anomalous process indicates that the DCA may be suitable as a general purpose anomaly detector. The addition of PAMPs does not significantly increase the detection of the 'anomalous' nmap ($p > 0.05$), but combined with a higher safe signal weight, significantly lowers the % mature antigen value of the normal processes. This implies that in future experiments a much higher level of safe signal can be used without reducing the detection of the misbehaving process. This in turn may reduce the amount of false positives in comparison with previous AIS intrusion detection systems.

TABLE VIII
THREE SIGNALS WITH -2 WEIGHT (EXPT 3)

| Process Name | Mean % mAg | Standard Deviation |
|---|---|---|
| bash | 0.107 | 0.115 |
| sshd jqg@pts/8 | 0.271 | 0.077 |
| nmap -sP | 0.829 | 0.083 |
| X forward agt | 0.172 | 0.203 |
| scp blufire.ai | 0.176 | 0.196 |

TABLE IX
FOUR SIGNALS (EXPT 4)

| Process Name | Mean % mAg | Standard Deviation |
|---|---|---|
| bash | 0.103 | 0.054 |
| sshd jqg@pts/8 | 0.423 | 0.121 |
| nmap -sP | 0.796 | 0.139 |
| X forward agt | 0.319 | 0.160 |
| scp blufire.ai | 0.348 | 0.227 |

TABLE X
COMPARISON OF ANOMALOUS NMAP AND NORMAL SECURE COPY (SCP) BASED ON DISTANCE OF MEAN % MATURE ANTIGEN

| Experiments | distance of means | p value |
|---|---|---|
| 2 signals (-1 ss) | 0.465 | <0.001 |
| 3 signals (-1 ss) | 0.482 | <0.001 |
| 3 signals (-2 ss) | 0.652 | <0.001 |
| 4 signals (-2 ss) | 0.448 | <0.001 |

It is also interesting to note that the inclusion of inflammatory signals in experiment 4 produces an increase in the detection of the normal process (0.348 %mAg) when compared to experiment 3 (0.176 %mAg). On closer inspection of the output data we discovered that fewer antigen are presented per DC (0.25 antigen per DC) versus 0.42 antigen per DC in all three previous experiments. Two implications are evident from these results: First, DCs spend a shorter time in the tissue as the CSM threshold is exceeded at twice the rate; second, as the DC spend less time in the tissue, less antigen is collected. This is analogous to the effects of inflammation in the human immune system.

## VII. CONCLUSIONS AND FURTHER WORK

In this paper we have demonstrated the use of a Dendritic Cell inspired Algorithm (DCA) on two datasets. The results from the Wisconsin Breast Cancer experiments have validated the use of the algorithm within the libtissue framework. The promising results shown in the port scan experiments imply that the DCA plus the libtissue framework can be used for the purpose of anomaly detection under real-time conditions.

Future work includes the incorporation of the replay client outlined in Section III. This client will allow us to collect more data of our choice, and to study the behaviour of the DCA in detail. Sensitivity analysis performed using the replay client will give us greater understanding of the DCA, necessary before any further development of the algorithm. A suitable method of comparison with other anomaly detection techniques is also under investigation. Ultimately, we would like to apply this data to real-time system monitoring to detect exploits, botnets or scanning worms. Results from this will assist in deriving an answer to the original question[4]: is Danger Theory the missing link between AIS and IDS?

## VIII. ACKNOWLEDGMENTS

This project is supported by the EPSRC (GR/S47809/01), UWE, Hewlett Packard Labs, Bristol, and the Firestorm IDS team. Thanks to Jim Greensmith, Markus Hammonds and Dr Jon Timmis for their comments.